\documentclass[sigconf]{acmart} 
\AtBeginDocument{%
  }

\setcopyright{acmlicensed}
\copyrightyear{2018}
\acmYear{2018}
\acmDOI{XXXXXXX.XXXXXXX}

\acmConference[Conference acronym 'XX]{Make sure to enter the correct
  conference title from your rights confirmation emai}{June 03--05,
  2018}{Woodstock, NY}
\acmISBN{978-1-4503-XXXX-X/18/06}

\usepackage{algorithm}
\usepackage{algpseudocode}
\usepackage{adjustbox}
\usepackage[utf8]{inputenc}
\usepackage{url} 
\usepackage{hyperref} 
\usepackage{enumitem} 
\usepackage{booktabs}
\usepackage{graphicx}
\usepackage{lscape}
\usepackage{wrapfig}
\usepackage{graphicx}
\usepackage{caption}
\usepackage{subcaption}
\usepackage{array}
\usepackage{tabularx}
\usepackage{wrapfig}
\usepackage{booktabs}
\usepackage{multirow}
\usepackage{soul}
\usepackage{colortbl}

\begin{document}

\title{\textit{AnxietyFaceTrack:} A Smartphone-Based Non-Intrusive Approach for Detecting Social Anxiety Using Facial Features}

\author{Nilesh Kumar Sahu}
\email{nilesh21@iiserb.ac.in}
\orcid{0000-0003-1675-7270}
\affiliation{%
  \institution{IISER Bhopal}
  \streetaddress{Bhauri}
  \city{Bhopal}
  \state{}
  \country{India}
  \postcode{462066}
}

\author{Snehil Gupta}
\email{snehil.psy@aiimsbhopal.edu.in}
\affiliation{%
  \institution{AIIMS Bhopal}
  \streetaddress{Saket Nagar}
  \city{Bhopal}
  \country{India}}

\author{Haroon R. Lone}
\email{haroon@iiserb.ac.in}
\affiliation{%
  \institution{IISER Bhopal}
  \streetaddress{Bhauri}
  \city{Bhopal}
  \state{ }
  \country{India}
  \postcode{462066}
}

\begin{abstract}
Social Anxiety Disorder (SAD) is a widespread mental health condition, yet its lack of objective markers hinders timely detection and intervention. While previous research has focused on behavioral and non-verbal markers of SAD in structured activities (e.g., speeches or interviews), these settings fail to replicate real-world, unstructured social interactions fully. Identifying non-verbal markers in naturalistic, unstaged environments is essential for developing ubiquitous and non-intrusive monitoring solutions.
To address this gap, we present \textit{AnxietyFaceTrack}, a study leveraging facial video analysis to detect anxiety in unstaged social settings. A cohort of 91 participants engaged in a social setting with unfamiliar individuals and their facial videos were recorded using a low-cost smartphone camera. We examined facial features, including eye movements, head position, facial landmarks, and facial action units, and used self-reported survey data to establish ground truth for multiclass (anxious, neutral, non-anxious) and binary (e.g., anxious vs. neutral) classifications.
Our results demonstrate that a Random Forest classifier trained on the top 20\% of features achieved the highest accuracy of 91.0\% for multiclass classification and an average accuracy of 92.33\% across binary classifications. Notably, head position and facial landmarks yielded the best performance for individual facial regions, achieving 85.0\% and 88.0\% accuracy, respectively, in multiclass classification, and 89.66\% and 91.0\% accuracy, respectively, across binary classifications. Post-hoc analysis identified head rotation (x-axis), facial edge features, and eye landmarks as key contributors to detecting anxiety.
This study introduces a non-intrusive, cost-effective solution that can be seamlessly integrated into everyday smartphones for continuous anxiety monitoring, offering a promising pathway for early detection and intervention.
\end{abstract}


\begin{teaserfigure}
    \centering
    \includegraphics[width=1\linewidth]{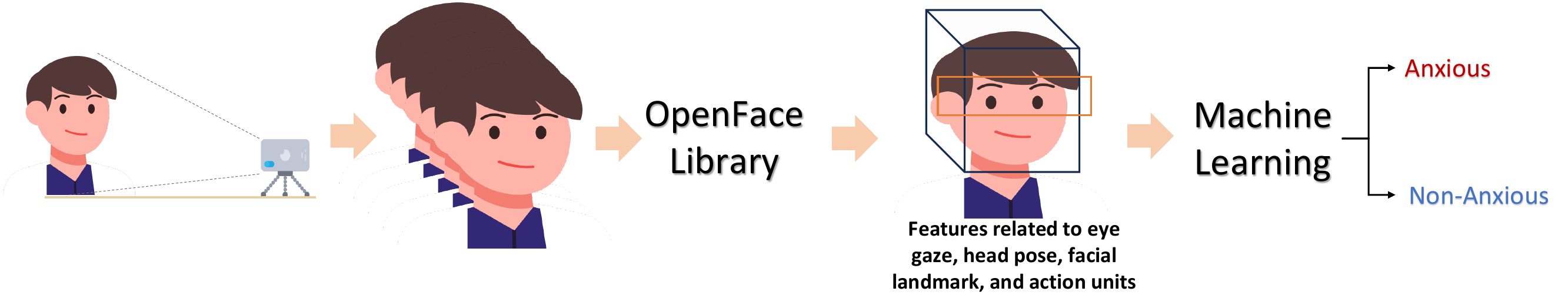}
    \caption{Anxiety detection framework of \textit{AnxietyFaceTrack}.}
    \Description{}
    \label{fig:framework}
\end{teaserfigure}

\received{20 February 2007}
\received[revised]{12 March 2009}
\received[accepted]{5 June 2009}

\maketitle

\section{Introduction} \label{sec:intro}

Social Anxiety Disorder (SAD) is characterized by excessive fear and worry and can manifest in various ways, including physical symptoms such as a racing heartbeat and sweating, mental symptoms like restlessness and pervasive fear, and behavioral symptoms such as avoidance of social activities \cite{szuhany2022anxiety}. One in every eight individuals experiences a mental disorder, with anxiety being the most prevalent \cite{Mentaldi93:online}. According to the Global Burden of Disease 2019, anxiety disorder ranks second leading mental health-related contributor to disability-adjusted life-years (DALYs) and years lived with disability (YLDs) globally \cite{xiong2022trends}. Furthermore, the COVID-19 pandemic led to a significant 26\% increase in the number of individuals suffering from anxiety disorders \cite{Mentaldi93:online}.  

Currently, SAD diagnosis relies upon traditional methods due to the absence of any reliable, objective markers (or measures) of anxiety disorder. The traditional method includes clinical interviews and clinically validated retrospective self-reported questionnaires. However, these traditional methods have limitations, such as clinical interviews are prone to human bias and rely on the subject self-motivation to attend the interviews that require multiple sessions, while self-reporting relies on the subject willingness to convey their behaviors and is prone to recall bias. Thus, given the high prevalence of SAD, there is a need for automated, reliable measures that are not susceptible to human bias.

Ongoing research on mental disorder detection has explored unobtrusive objective markers using methods such as wearable sensors, speech analysis, and mobile phone data \cite{rashid2020predicting, salekin2018weakly}. However, the findings remain inconclusive. While facial images and videos have shown promise in detecting depression \cite{nepal2024moodcapture, ringeval2019avec}, there is a notable lack of studies focusing on the detection of SAD using these methods. Prior research indicates that individuals with SAD often exhibit behavioral and non-verbal cues \cite{gilboa2013more}, including restlessness \cite{Chand2023}, reduced motion \cite{Chand2023}, avoidance of eye contact \cite{schneier2011fear}, gaze fixation, slumped posture, and closed body language \cite{weeks2011exploring}.

Existing research on detecting mental disorders or emotional states from video predominantly relies on participants engaging in structured, anxiety-inducing tasks, such as delivering speeches \cite{harshit2024eyes}, introducing themselves \cite{shafique2022towards, giannakakis2017stress}, or watching stressful videos \cite{giannakakis2017stress}. These studies often depend on high-end recording equipment \cite{gavrilescu2019predicting}, such as RGB-Depth cameras \cite{horigome2020evaluating}, which may not be feasible for widespread use. However, there is a significant gap in exploring whether anxiety can be detected in natural, unstaged social interactions using low-cost video cameras without requiring participants to engage in specific activities. Addressing this gap is critical for making anxiety detection more accessible and applicable to real-world scenarios.

To address the gap in detecting SAD in natural social settings without requiring participants to perform specific activities, we developed \textit{AnxietyFaceTrack}. It enables the observation of natural expressions of social anxiety without the influence of artificial tasks or prompts. In the study, we invited participants, unfamiliar with one another, to sit in trios in a simulated social scenario. Each participant was positioned to face unknown individuals seated in front, left, and right, replicating real-life situations of encountering strangers. During the 2-minute session, participants were recorded using a dedicated low-cost smartphone camera (approximately \$81), which captured their upper body, including the face, shoulders, and chest. This design ensures accessibility and ecological validity, providing valuable insights into behavioral markers of SAD in naturalistic contexts.

A total of 91 university students participated in the study. Behavioral features, such as head position, gaze direction, and facial expressions, were extracted from the recorded videos and used to train classification models for anxiety detection. We developed a multiclass classification model to categorize participants as anxious, neutral, or non-anxious, using self-reported ground truth labels provided by the participants. This multiclass approach was adopted to account for neutral behaviors, which were observed among both SAD and non-SAD participants, and to reduce potential bias in the classification. Also, we evaluated binary classification performance as shown in Figure \ref{fig:framework} by excluding one category to focus on the distinction between two class behaviors. For example, in the ``anxious versus non-anxious'' model, only participants labeled as anxious or non-anxious were included, and the neutral was dropped.
 
The key contributions of our work are as follows:
\vspace{-2pt}
\begin{enumerate} 
    \item We present a non-intrusive approach for detecting anxiety in normal settings using a low-cost smartphone camera that can integrated into daily use smartphones for continuous monitoring of anxiety, thus prompting early interventions.
    \item We evaluated several machine learning and deep learning models for anxiety detection using facial features. We tested these models on 669 facial features and their subsets. Our results show that the Random Forest model outperformed others in nearly all classification metrics for both multiclass (Accuracy- 91\%, F1 score- 0.90, AUC- 0.98) and binary classification (Anxious vs. Neutral: Accuracy- 92\%, F1 score- 0.91, AUC- 0.98; Anxious vs. Non-Anxious: Accuracy- 92\%, F1 score- 0.89, AUC- 0.98; Neutral vs. Non-Anxious: Accuracy- 93\%, F1 score- 0.94, AUC- 0.98).
    \item We identified key features that helped the Random Forest model correctly identify anxious participants and other classes. For example, larger head rotation along the X-axis, face edge features (such as jawline points), and eye landmarks were important for anxiety detection in \textit{AnxietyFaceTrack}.
    \item We also analyzed the model’s bias and found that it performed better for females. This may be because females generally show higher facial expressions than males \cite{parkins2012gender, fischer2015drives, kring1998sex}.
\end{enumerate}

\textit{AnxietyFaceTrack} contributes to affective computing, showcasing the use case of facial features captured through videos for anxiety detection. The use of low-cost smartphone cameras and machine learning models using facial cues might offer a practical solution for continuously monitoring mental disorders, thus reducing the treatment gap and prompting early interventions. Furthermore,  our findings can serve as a baseline for future research conducted in controlled or uncontrolled settings for anxiety detection using facial features.

The paper is structured as follows: Section \ref{section: RW} presents related work on anxiety detection and studies that use videos for mental disorders. Section \ref{sec:method} explains our \textit{AnxietyFaceTrack} study, participant demographics, ground truth, and the analysis methods used for anxiety detection. Section \ref{section: results} discusses our results and the ablation study conducted to draw inferences about anxiety detection. Section \ref{section: feature_imp} presents the important features that influenced the detection of anxiety, while Section \ref{section: bias} examines the bias in the trained models. Section \ref{section: discuss} discusses the study’s findings, implications, and limitations, while Section \ref{section: conclusion} provides the conclusion.


\section{Related Works} \label{section: RW}

\subsection{Non-verbal Cues of Anxiety Disorder}
Over the last five decades, researchers have studied nonverbal communication in mental disorders \cite{waxer1977nonverbal, argyle1978non, perez2003nonverbal,foley2010nonverbal, schneier2011fear, gilboa2013more, weeks2019fear, asher2020out, shatz2024nonverbal}. These studies have found that nonverbal cues can play a significant role in diagnosing mental disorders and contribute to therapeutic processes. For instance, nonverbal signs such as a patient’s appearance, behavior, and eye contact are routinely assessed during psychiatric evaluations as part of the mental status examination \cite{foley2010nonverbal}.

Existing research has analyzed video recordings of individuals with anxiety disorders in various scenarios, such as therapy sessions, task performance, video watching, dyadic conversations, etc. One of the earliest studies on this topic was conducted by Waxer \cite{waxer1977nonverbal}, who analyzed the nonverbal cues of individuals with anxiety. In this study, anxious and non-anxious participants (20 participants: 5 anxious males, 5 non-anxious males, 5 anxious females, and 5 non-anxious females) were videotaped at different times during the admission period. One-minute silent session videos were then shared with 46 senior psychologists. They rated ten behavior cue areas on a 10-point scale ranging from ``not anxious at all'' to ``highly anxious'' and described how these features conveyed anxiety. The ten behavior cue areas were the forehead, eyebrows, eyelids, eyes, mouth, head angle, shoulder posture, arm position, torso position, and hands. Further, using Linear regression analysis, hands, eyes, mouth, and torso were identified as a key nonverbal indicator of anxiety.

A major focus of recent studies has been on gaze behavior and its relationship with social anxiety. Schneier et al. \cite{schneier2011fear} explored gaze avoidance in individuals with generalized social anxiety disorder, healthy controls, and undergraduate students. Their findings indicate that avoiding eye contact is associated with social anxiety. Similarly, Weeks et al. \cite{weeks2011exploring, weeks2019fear} conducted multiple studies on behavioral submissiveness in social anxiety. In one study, participants engaged in a role-play task with unfamiliar individuals (confederates), revealing that body collapse and gaze avoidance are linked to social anxiety \cite{weeks2011exploring}. In another study, Weeks et al. used eye-tracking systems to examine participants watching positive and negative video clips, further identifying gaze avoidance as a prominent marker of SAD \cite{weeks2019fear}.

Nonverbal synchrony is another area of investigation in SAD. Asher et al. \cite{asher2020out} analyzed dyadic conversations between individuals with SAD and non-anxious individuals, finding impaired nonverbal synchrony among those with SAD. Similarly, Shatz et al. \cite{shatz2024nonverbal} examined nonverbal synchrony during diagnostic interviews, showing that individuals with SAD displayed lower levels of synchrony and reduced pacing compared to non-anxious counterparts. An in-depth review by Gilboa et al. \cite{gilboa2013more} provides a comprehensive understanding of nonverbal social cues in SAD, synthesizing findings from various studies and emphasizing the role of nonverbal behaviors in the disorder. The findings from these studies underscore the role of nonverbal cues, such as gaze behavior and body posture in understanding and diagnosing SAD. These studies relied on recorded videos and human inference, thus highlighting the need for an automated tool.

\subsection{Visual Features of Mental Disorders}
To the best of our knowledge, Cohn et al. \cite{cohn2009detecting} were the first to explore the use of automated visual features from videos for research in mental health detection. They recorded interviews between clinically depressive participants and interviewers. The study used manual and automated facial analysis coding systems (FACS) as feature inputs for machine learning. They achieved accuracies of 88\% with manual features and 79\% with automated features in depression detection. Furthermore, `\textit{AVEC 2011 – The First International Audio/Visual Emotion Challenge}' introduced automated visual features, calculated using dense local appearance descriptors, for affective computing \cite{schuller2011avec}. This was done through a workshop challenge that provided an open dataset to the research community. Later, the development of OpenFace\footnote{\url{https://cmusatyalab.github.io/openface/}}, based on FaceNet \cite{schroff2015facenet}, an advanced deep learning model, offered a unified system for detecting facial features. Later, the updated version, OpenFace 2.0 \cite{baltruvsaitis2016openface}, emerged as a state-of-the-art computer vision toolkit. It enabled researchers to analyze facial behavior and study nonverbal communication without requiring comprehensive programming knowledge.

Most studies that use visual features for mental health detection have focused on depression and stress \cite{cohn2009detecting, schuller2011avec, valstar2014avec, ringeval2019avec, ringeval2017avec, valstar2016avec, valstar2013avec, wang2021multimodal, gavrilescu2019predicting, grimm2022phq, giannakakis2017stress, sun2022estimating}, with limited attention given to anxiety \cite{wang2021multimodal, gavrilescu2019predicting, grimm2022phq, mo2024multimodal, giannakakis2017stress} and even less to SAD \cite{harshit2024eyes, shafique2022towards}. Giannakakis et al. \cite{giannakakis2017stress} used an open-source model to detect the face region of interest and applied Active Appearance Models (AAM) for emotion recognition and facial expression analysis. In their study, participants were recorded with a video camera with extra lighting while undergoing three experimental phases: a social exposure phase, an emotion recall phase, and a stress/mental task phase. The computed features were then used to detect emotional states related to stress and anxiety. They achieved an average accuracy of 87.72\% in stress detection across these phases. Similarly, Sun et al. \cite{sun2022estimating} utilized visual features for remote stress detection. Participants attended an online meeting and self-reported their stress levels on a scale of 1 to 10, which served as the ground truth for a binary stress classifier. The study reported an accuracy of 70.00\% using motion features (eye and head movements) and 73.75\% using facial expressions (action units). In another study, Grimm et al. \cite{grimm2022phq} analyzed participants’ videos captured while they answered open-ended questions. A classifier was trained using GAD-7 scores as ground truth, achieving an area under the curve (AUC) score of 0.71 for the binary classification of anxiety characteristics. Similarly, Gavrilescu et al. \cite{gavrilescu2019predicting} predicted depression, anxiety, and stress using videos captured with high-end cameras while participants watched emotion-inducing clips. They achieved accuracies of 87.2\% for depression, 77.9\% for anxiety, and 90.2\% for stress.

Existing studies on SAD have predominantly focused on eye gaze. In these studies, participants typically complete a performance task involving interviews or the Trier Social Stress Test (TSST). For example, Shafique et al. \cite{shafique2022towards} used participants' eye gaze data captured during a 5-minute general conversation with an examiner, covering topics such as introductions, support, and conflict. Their method achieved an accuracy of 80\% in detecting the severity of SAD. In another study, Harshit et al. \cite{harshit2024eyes} analyzed participant's eye gaze while they performed a speech task as part of the TSST. Using an autoencoder, they extracted latent feature representations (deep features) from the eye gaze data, which was used as features for machine learning models, and achieved 100\% accuracy in detecting participants' anxiety.

In summary, most existing studies focus on interview-based videos or externally induced anxiety tasks, limiting their relevance to real-world, everyday scenarios. Additionally, non-verbal cues, such as facial expressions and head movements, remain underexplored in detecting SAD. To address these gaps, we designed a study set in a social environment where participants were surrounded by unfamiliar individuals and instructed to remain idle without engaging in any activity. Using a low-cost smartphone camera, we recorded videos of the participants' faces, extracted facial features, and analyzed them for insights.

\section{Methodology} \label{sec:method}
In this section, we present the study design of \textit{AnxietyFaceTrack}, participants' demographics, ground truth collection, feature extraction, and classification models.

\subsection{Study Design} 
Participants were recruited from the home institute through an email advertisement, following approval from the Institutional Review Board.
 A dedicated email was sent to the student community with information related to the study and the Google form to fill out for the interested participants. The participants filled out their demographic information, such as age, gender, current educational program, location of home residence, and preferred time slot. Additionally, the participants' email and phone numbers were collected so the research assistant (RA) could contact them on the study day.  

The day before the study, the RA sent an email to the participants to confirm their availability for a specific time slot. Further, a text message was sent one hour before the study session, confirming the location and time for participation. Three participants were invited to the lab for each study session. The RA ensured that the three participants were unfamiliar and did not know each other. The purpose of inviting three unfamiliar participants was to create a socially anxious situation during the study.

\begin{figure}
    \centering
    \includegraphics[width=1\linewidth]{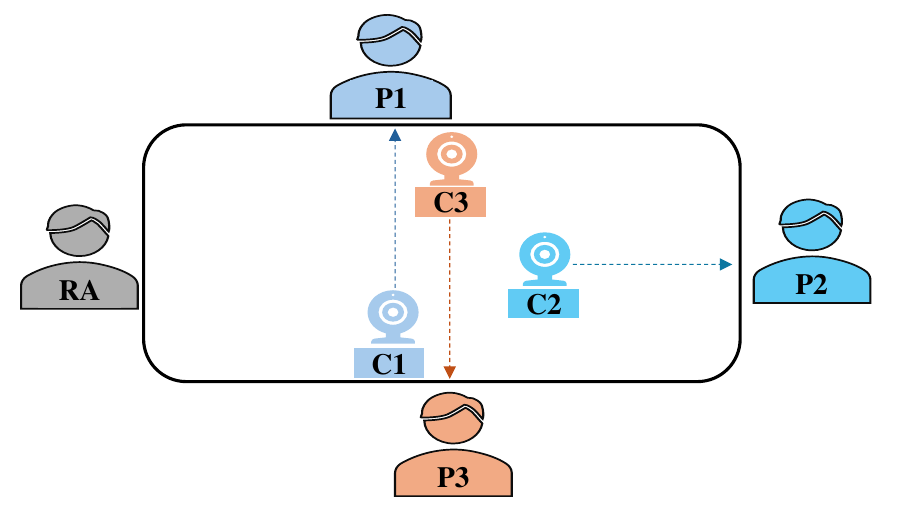}
    \caption{The study setup shows participants' sitting positions and camera positions. RA refers to Research Assistant, and labels P1, P2, and P3 refer to participants 1, 2, and 3, respectively. Labels C1, C2, and C3 refer to smartphone cameras 1, 2, and 3, respectively.}
    \Description{}
    \label{fig:sitting_position}
\end{figure}

Upon arrival, the participants were seated around a rectangular table with rounded edges. Each session involved three participants and a RA. The participants were labeled as P1, P2, and P3 for each study session. The seating arrangement is shown in Figure \ref{fig:sitting_position}: P1 was seated to the left of the RA, P2 was directly opposite the RA, and P3 was to the right of the RA, where P1 and P3 faced each other, while P2 faced the RA. This arrangement ensured that each participant faced an unfamiliar person, creating a socially anxious scenario. The RA explained the study to the participants and distributed the Participant Information Sheet (PIS) and the Informed Consent Form (ICF). After collecting the signed consent forms, the RA obtained permission to start camera recordings. Three individual smartphones, labeled C1, C2, and C3, with 13-megapixel back cameras were used to record P1, P2, and P3, respectively (see Figure \ref{fig:sitting_position}).  The ``Background Video Recorder'' app was selected for its ability to record video even with the screen off, which was not possible with the smartphone’s default camera app. The app was configured to use the highest possible sampling rate of 30 frames per second (FPS).

The RA instructed participants to remain seated for two minutes without interacting with others. Participants were free to look around but remained idle. After two minutes, the RA concluded the session, distributed a self-reported survey (discussed later), and recorded the session’s start and end times. Finally, the RA thanked the participants and provided refreshments as a token of appreciation for their time and participation.

\subsection{Demographics}
The participants in our study are the student population of the author's institute. Most participants were male (\#58, 63.74\%), followed by female (\#33, 36.26\%). In terms of education status, 71.43\% (\#65) were undergraduate students, while the remaining 28.57\% (\#26) were graduate students. Regarding home location, 76.92\% (\#70) were from urban areas, and 23.08\% (\#21) were from rural areas. A detailed breakdown of the participants' demographic information is provided in Table \ref{tab: demographic}.

\begin{table}[]
\centering
\small
\caption{Participants' demographics. SR: Self-reported}
\label{tab: demographic}
\begin{tabular}{@{}lcccclcc@{}}
\toprule
\textbf{Category} & \multicolumn{1}{l}{\textbf{Count}} & \multicolumn{1}{l}{\textbf{Percentage}} & \multicolumn{2}{c}{\textbf{Age}} & & \multicolumn{2}{c}{\textbf{SR anxiety}} \\ \cmidrule{4-5} \cmidrule{7-8}
 & \multicolumn{1}{c}{(\#)} & \multicolumn{1}{c}{(\%)} & \multicolumn{1}{c}{$\mu$} & \multicolumn{1}{c}{$\sigma$} & & \multicolumn{1}{c}{$\mu$} & \multicolumn{1}{c}{$\sigma$} \\ \midrule
\multicolumn{8}{l}{\cellcolor[HTML]{EFEFEF}\textit{\textbf{Gender}}} \\
Female & 33 & 36.26 & 20.94 & 2.73  && 3.27& 1.13\\
Male & 58 & 63.74 & 20.59 & 2.13  && 3.29& 0.97 \\

\multicolumn{8}{l}{\cellcolor[HTML]{EFEFEF}\textit{\textbf{Education}}} \\
Graduate & 26 & 28.57 & 23.69 & 2.04  && 3.42& 1.06 \\
Undergraduate & 65 & 71.43 & 19.52 & 1.06  && 3.23& 1.01 \\

\multicolumn{8}{l}{\cellcolor[HTML]{EFEFEF}\textit{\textbf{Home Location}}} \\
Rural & 21 & 23.08 & 20.95 & 2.56  && 3.86& 0.96 \\
Urban & 70 & 76.92 & 20.64 & 2.3  && 3.11& 0.99 \\  \midrule
\textbf{\textit{Total}} & 91 & 100 & 20.71 & 2.35  && 3.28 & 1.02 \\ \bottomrule
\end{tabular}
\end{table}

\subsection{Ground Truth}
\begin{figure}
    \centering
    \includegraphics[width=0.8\linewidth]{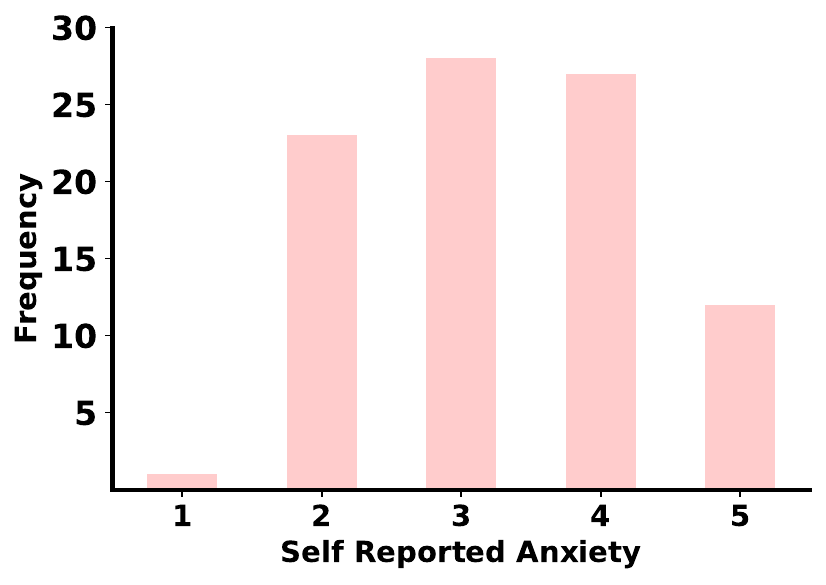}
    \caption{Distribution of participants' self-reported anxiety}
    \Description{}
    \label{fig:histogram_baseline}
\end{figure}

The self-reported survey collected during the study was used as the ground truth. The survey included a single question asking participants about their anxiety levels during the studys session (i.e., sitting idle for 2 minutes). Participants rated their anxiety on a Likert scale from 1 to 5, where: 1: Very nervous, 2: Somewhat nervous, 3: Neither relaxed nor nervous, 4: Somewhat relaxed, and 5: Very relaxed. The distribution of self-reported anxiety ratings is shown in Figure \ref{fig:histogram_baseline}.

For the ground truth, participants who rated their anxiety as 1 or 2 were labeled as \textit{anxious}, while those who rated their anxiety as 4 or 5 were labeled as \textit{non-anxious}. A considerable number of participants rated their anxiety as 3, so instead of grouping these participants into either the anxious or non-anxious categories, they were labeled as \textit{neutral}. 

\subsection{Feature Extraction} \label{section: Data Preparation}

During data processing, the recorded videos had an average duration of 124.21 seconds and a mean sampling rate of 20.02 FPS. To ensure uniformity, we selected the first 90 seconds of each video for analysis, as participants generally exhibited reduced anxiety over time. Videos shorter than 90 seconds were excluded, while longer videos were trimmed to the initial 90 seconds. Data from six participants were lost due to issues such as delayed recording initiation by the research assistant or technical problems, including incomplete recordings caused by memory errors or full storage. Ultimately, data from 85 participants were retained for further analysis. This approach minimized data loss while maintaining a consistent dataset.

To extract features from participant's videos, we used the open-source OpenFace\footnote{\url{https://github.com/TadasBaltrusaitis/OpenFace}} Python toolkit \cite{baltruvsaitis2016openface}. OpenFace is a well-validated Python-based tool for facial analysis tasks and has been widely used in various behavioral studies, including depression and anxiety detection. The toolkit processes video inputs and generates time-series data consisting of 714 features. In our analysis, we excluded metadata information features (\#5) and rigidity features (\#40) due to limited relevance in existing literature and lack of interpretability. The remaining 669 features used in the analysis are described in Table \ref{tab:openface_features}.

Following methodologies proposed by Bhatti et al. \cite{bhatti2024attx} and Schmidt et al. \cite{schmidt2018introducing}, the data was prepared for classification through two key steps: (i) \textit{Chunking:} The 669 OpenFace features were segmented using non-overlapping windows of 10 seconds to create data chunks. (ii) \textit{Flattening:} For each chunk, the mean values of all features were computed along the time dimension, resulting in a data sample of size $1 \times  669$. 

This process produced a dataset of 1,173 samples. Ground truth labels were then associated with the prepared dataset, resulting in the following class distribution: anxious (314 samples), neutral (384 samples), and non-anxious (475 samples). Additional details on sample distributions are provided in Table \ref{tab:samples}.

\begin{table*}[htbp]
\centering
\caption{Summary of OpenFace output features \cite{OpenFaceOutputFormat}.}
\label{tab:openface_features}
\begin{tabular}{cccp{0.70\textwidth}}  
\toprule
\textbf{Feature} & \textbf{Feature Set} & \textbf{\#} & \textbf{Description} \\ \midrule

\multirow{3}{*}{Eye} 
    & \multirow{3}{*}{\shortstack{gaze \\ 2D landmarks \\ 3D landmarks}}  & \multirow{3}{*}{\shortstack{8 \\ 112 \\ 168}} & Gaze represents the direction in which an individual is looking, using 3D vector world coordinates for both the left (\textit{gaze\_0\_x, gaze\_0\_y, gaze\_0\_z}) and right (\textit{gaze\_1\_x, gaze\_1\_y, gaze\_1\_z}) eyes. It also includes the gaze direction angle (\textit{gaze\_angle\_x, gaze\_angle\_y}), averaged for both eyes, indicating whether a person looks left-right or up-down. Eye landmarks (pupil and eyelids) represent the position of landmarks around the eye region in both 2D (\textit{eye\_lmk\_x\_0, eye\_lmk\_x\_1,... eye\_lmk\_x55, eye\_lmk\_y\_1,... eye\_lmk\_y\_55}) and 3D (eye\_lmk\_X\_0, eye\_lmk\_X\_1,... eye\_lmk\_X55, eye\_lmk\_Y\_0,... eye\_lmk\_Z\_55) coordinates.
    \\ \midrule
    
\multirow{2}{*}{\shortstack{Head \\ Pose}} 
    & \multirow{2}{*}{\shortstack{location \\ rotation}}  & \multirow{2}{*}{\shortstack{3 \\ 3}}  & Pose location (\textit{pose\_Tx, pose\_Ty, pose\_Tz}) represents the position of the head relative to the camera along the X, Y, and Z axes, while the pose angle (\textit{pose\_Rx, pose\_Ry, pose\_Rz}) represents the rotation of the head along these axes. The rotations are referred to as pitch (X axis), yaw (Y axis), and roll (Z axis). \\ \midrule
    
\multirow{2}{*}{\shortstack{Face \\ Landmarks}} 
    & \multirow{2}{*}{\shortstack{ 2D landmarks\\ 3D landmarks}} & \multirow{2}{*}{\shortstack{ 136 \\ 204}} & Face landmarks represent 68 key positions on the face. These positions include the jawline (i.e., face edge) with 17 points (0 to 16), eyebrows for the left and right eyes with 10 points (17 to 26), the nose bridge and tip with 9 points (27 to 35), eyes (left and right) with 12 points (36 to 47), and the mouth, consisting of the upper lip, lower lip, and corners, with 20 points (48 to 67). The landmarks are present in both 2D (\textit{x\_0, x\_1, ... x\_66, x\_67, y\_0,...y\_67}) and 3D (X\_0, ... X\_67, Y\_0,...Y\_67, Z\_0,...Z\_67) coordinates. 
    \\ \midrule

\multirow{2}{*}{\shortstack{Facial Action \\ Units (AUs)}} 
    & \multirow{2}{*}{\shortstack{ intensity \\ presence}}  & \multirow{2}{*}{\shortstack{ 17 \\ 18}}  & AUs represent facial expressions using the Facial Action Coding System based on muscle movements. AUs are described in terms of presence (AU\#\_c) and intensity (AU\#\_r). Presence indicates whether the AU is visible on the face, while intensity refers to how strong the AU is, measured on a scale of 1 to 5. \\     \bottomrule
\end{tabular}
\end{table*}

\begin{table}[]
\centering
\small
\caption{Distribution of ``Anxious'', ``Neutral'', and ``Non-anxious'' samples in the prepared dataset.}
\label{tab:samples}
\begin{tabular}{@{}lccc@{}} 
\toprule
 & \textbf{Anxious (\#)} & \textbf{Neutral (\#)} & \textbf{Non-anxious (\#)} \\ \midrule
\multicolumn{4}{l}{\cellcolor[HTML]{EFEFEF}\textit{\textbf{Gender}}} \\
Female & 128 & 146 & 166 \\
Male & 186 & 238 & 309 \\

\multicolumn{4}{l}{\cellcolor[HTML]{EFEFEF}\textit{\textbf{Education}}} \\
Graduate & 69 & 114 & 139 \\
Undergraduate & 245 & 270 & 336 \\ 

\multicolumn{4}{l}{\cellcolor[HTML]{EFEFEF}\textit{\textbf{Home Location}}} \\
Rural & 29 & 65 & 163 \\
Urban & 285 & 312 & 319 \\ \midrule \midrule
Total & 314 & 384 & 475 \\ \bottomrule
\end{tabular}
\end{table}

\subsection{Anxiety Classification}

\subsubsection{Machine Learning (ML)}
We used various classification models, each leveraging distinct strengths and methodologies, to identify the most effective model for anxiety classification using facial features. Logistic Regression (LR) \cite{hosmer2013applied} was utilized for its simplicity and effectiveness in binary classification tasks. K-Nearest Neighbors (KNN) \cite{zhang2016introduction} was included to assess the potential of proximity-based classification. Support Vector Machines (SVM) \cite{brereton2010support} were selected for their capability to handle high-dimensional feature spaces effectively. Decision Trees (DT) \cite{song2015decision} offered interpretability, allowing for a better understanding of feature contributions, while Random Forests (RF) \cite{breiman2001random} provided robustness against noise and the ability to capture complex feature interactions.

\subsubsection{Deep Learning (DL)}
We also applied deep learning models to the extracted facial features for anxiety classification. Specifically, we used a multilayer perceptron (MLP) \cite{alnuaim2022human} and a one-dimensional convolutional neural network (1D CNN) \cite{kiranyaz20211d}. The reason for using deep learning models was their ability to learn complex patterns. However, we did not use other advanced deep learning models, as these require large amounts of data for training, and we had a limited dataset. Further, these models are computationally expensive, thus limiting the use case of our study. The MLP model used in this paper had an input layer connected to two dense layers with 64 and 32 neurons, followed by an output layer. Similarly, the 1D CNN had four convolutional layers with 64, 128, 256, and 128 neurons. The Adam optimizer was used for both models, with categorical cross-entropy as the loss function.

\subsubsection{Ablation Studies}
We conducted three ablation studies to analyze the performance and impact of different features and classification tasks:
(i) \textit{Ablation Study 1:} This study identified the most impactful feature for anxiety classification using Random Forest feature importance.
(ii) \textit{Ablation Study 2:} This study assessed the effectiveness of various feature categories (see Table \ref{tab:openface_features}) to determine which category contributed most significantly to the classification.
(iii) \textit{Ablation Study 3:} This study evaluated the classification model's performance on three binary classification tasks: anxious vs. non-anxious, anxious vs. neutral, and neutral vs. non-anxious.

\subsubsection{Evaluation}
To evaluate the trained classification models, we used a 5-fold cross-validation approach \cite{kohavi1995study}. This approach ensures that the model is trained on different subsets of the data in each iteration and tested on unseen data, providing a more reliable and robust evaluation compared to a single train-test strategy. Further, to assess classification performance, we used evaluation metrics, including accuracy, precision, recall, F1-score, and area under the curve (AUC) \cite{sokolova2009systematic}. Accuracy, precision, and recall range from 0 to 100\%, while the F1-score and AUC range from 0 to 1. Higher values indicate better performance. These metrics were computed for each fold and then averaged across all five folds. 


\section{Results}  \label{section: results}
In this section, we present the outcomes of our analysis, including the predictive capabilities of the ML and DL classification models used for the multiclass problem using facial features. Additionally, we discuss the results of the ablation studies conducted in three different scenarios: (i) classification using different feature categories, (ii) classification using handcrafted features, and (iii) classification for binary classification problems.

\subsection{Classification Performance}
In our analysis, we used classical ML and DL classification models to evaluate the ability of \textit{AnxietyFaceTrack} to detect anxiety in a lab setting without introducing additional anxiety-provoking situations. Table \ref{tab:Multiclass classification results} presents the performance metrics for all the classification models used. Specifically, Table \ref{tab:Multiclass classification results} shows each class's average precision and recall to assess the model's performance for each class. In the case of ML, we found that KNN and RF performed well, while LR showed the poorest performance in terms of accuracy. Further inspection of the other metrics revealed that RF and KNN achieved almost identical average precision and recall. However, when focusing on the ``anxious'' label, RF outperformed KNN, achieving higher average precision for this label.

Moreover, Table \ref{tab:Multiclass classification results} provides additional insights. Firstly, RF outperformed DT across all metrics, suggesting that the single-tree structure of DT suffered from overfitting, whereas the ensemble approach of RF (using multiple trees) effectively handled high-dimensional data without overfitting. Secondly, RF outperformed LR on all metrics. This is likely because LR relies on linear decision boundaries, which failed to capture the complex and non-linear patterns in the data, while RF could handle these complexities. The overall best performance of RF in multiclass classification inspired us to conduct an ablation study to assess its effectiveness in binary classification scenarios (see Section \ref{section: AS_binary}), such as distinguishing between ``anxious'' and ``non-anxious'' participants.

In the case of DL models, the 1D CNN outperformed the MLP on most evaluation metrics. The 1D CNN achieved an average accuracy of 84\%, compared to 80\% for the MLP. Although these DL models performed better than most used ML models, they still lagged behind the RF and KNN. This could be due to the ability of machine learning models to learn effectively with smaller datasets, while deep learning models require larger amounts of data.

\textit{Given the overall best performance of RF across all evaluation metrics, we will now use the RF model in various ablation studies, which are discussed later in the paper.}



\begin{table}
    \centering
    \small
    \caption{Multiclass classification results averaged over five folds. 
    Abbreviations - Clf., Acc., Pr., and Re. refer to Classifier, Accuracy, Precision, and Recall, respectively. Metrics (Acc., Pr., and Re.) can be multiplied by 100 to represent percentages.}
    \label{tab:Multiclass classification results}
    \begin{tabular}{ccccccc} \toprule
         \textbf{Clf.}& \textbf{Acc.} & \textbf{Anxious}& \textbf{Neutral}& \textbf{Non-anx.}& \multirow{2}{*}{\shortstack{\textbf{F1} \\ \textbf{Score}}} & \textbf{AUC} \\
 & & \textbf{\textit{(Pr., Re.)}}& \textbf{\textit{(Pr., Re.)}}& \textbf{\textit{(Pr., Re.)}}&  &\\ \midrule
         LR & 0.65 & (0.64, 0.60) & (0.68, 0.69) & (0.65, 0.67) & 0.65 & 0.83 \\
         KNN & 0.86 & (0.81, 0.91) & (0.89, 0.84) & (0.89, 0.86) & 0.86 & 0.96 \\
         SVM & 0.73 & (0.68, 0.61) & (0.76, 0.76) & (0.73, 0.79) & 0.72 & 0.90 \\
         DT & 0.75 & (0.71, 0.70) & (0.75, 0.74) & (0.77, 0.79) & 0.74 & 0.81 \\
         RF & 0.88 & (0.86, 0.86) & (0.87, 0.88) & (0.90, 0.88) & 0.88 & 0.97 \\ \midrule \midrule
         MLP& 0.80 & (0.75, 0.72) & (0.82, 0.87) & (0.81, 0.80) & 0.80 & 0.94 \\
         1D CNN& 0.84 & (0.81, 0.81) & (0.87, 0.81) & (0.83, 0.87) & 0.83 & 0.95 \\ \bottomrule
    \end{tabular}
\end{table}

\begin{table}[]
\centering
\scriptsize
\caption{Multiclass classification results of Random Forest averaged over five folds. Abbreviations - Acc., Pr., Re., AS 1, and AS 2 refer to Accuracy, Precision, Recall, Ablation study 1, and Ablation study 2, respectively. Metrics (Acc., Pr., and Re.) can be multiplied by 100 to represent percentages.
}
\label{tab:3_class_classification_metric_RF}
\begin{tabular}{@{}ccccccccc@{}}
\toprule
&\multicolumn{1}{c}{\textbf{Feature}} & \textbf{Feature Set} & \textbf{\#} &  \textbf{Acc.}& \textbf{Pr.}& \textbf{Re.}& \textbf{F1}&\textbf{AUC}\\ \midrule

\multirow{2}{*}{\textbf{\textit{AS 1}}} & \multirow{2}{*}{ALL} & Top 10\% & 67& 0.90& 0.90& 0.90& 0.90&0.98 \\
& &  Top 20\%& 134& 0.91& 0.90& 0.91& 0.90&0.98\\ \midrule \midrule
 
\multirow{13}{*}{\textbf{\textit{AS 2}}} &\multirow{4}{*}{Eye} & gaze& 8 &   0.53& 0.52& 0.51& 0.52&0.72 \\
& & 2D landmarks & 112 &   0.70& 0.70& 0.69& 0.69&0.86 \\
& & 3D landmarks& 168 &   0.74& 0.73& 0.73& 0.73&0.9 \\ \cmidrule{3-3}
& & Combined& & 0.79& 0.79& 0.79& 0.79&0.93 \\ \cline{2-9}
 
& \multirow{3}{*}{Head Pose} & location & 3 &   0.81& 0.80& 0.80& 0.80&0.93 \\
& & rotation& 3 &   0.56& 0.56& 0.54& 0.55&0.74 \\ \cmidrule{3-3}
& & Combined& & 0.85& 0.85& 0.85& 0.85&0.96 \\ \cline{2-9}
 
& \multirow{3}{*}{Face Landmark} & 2D landmarks & 136 &   0.81& 0.81& 0.81& 0.81&0.94 \\
& & 3D landmarks& 204 &   0.88& 0.88& 0.87& 0.87&0.97 \\ \cmidrule{3-3}
& & Combined& & 0.87& 0.87& 0.87& 0.87&0.97 \\ \cline{2-9}
 
& \multirow{3}{*}{Facial Action Units} & intensity& 17 &   0.58& 0.58& 0.57& 0.57&0.78 \\
& & presence& 18 &   0.59& 0.6& 0.57& 0.58&0.76 \\ \cmidrule{3-3}
& & Combined& & 0.66& 0.67& 0.65& 0.65&0.84\\ \bottomrule
\end{tabular}
\end{table}

\subsection{Ablation Studies}
\subsubsection{Ablation Study 1}
In this analysis, we aimed to understand the role of feature importance in anxiety classification. We identified the most important features using the feature importance module from the Scikit-learn Random Forest implementation. We then trained and tested the model using k-fold cross-validation, incrementally selecting the top 10\%, 20\%, and up to 90\% of the features ranked by importance (see Table \ref{tab:3_class_classification_metric_RF}). Our findings revealed that using only the top 10\% of important features resulted in an accuracy of 90\%, which was 2\% higher than the accuracy achieved using all features. Increasing to the top 20\% of important features further improved accuracy to 91\%. Notably, the accuracy remained constant when using 30\% and 40\% of the features but began to decline after that, with a drop of just 1–2\%. This analysis demonstrates that using only the top 67 features (10\% of the total 669) achieves the highest accuracy, significantly reducing the model's complexity while maintaining good performance (see Table \ref{tab:3_class_classification_metric_RF}). This finding underscores the effectiveness of feature selection in optimizing classification models in anxiety detection.

\subsubsection{Ablation Study 2}
In this ablation study, we evaluated each feature set listed in Table \ref{tab:openface_features} to assess their ability in anxiety classification tasks. We selected RF as the classification model for this ablation study due to its superior performance compared to other models in previous analyses. The model was first trained and tested on individual feature sets and combinations of feature sets in individual facial regions, such as the combination of location and rotation in the head pose, etc. Table \ref{tab:3_class_classification_metric_RF} summarizes the results obtained from 5-fold cross-validation, with the classification metrics averaged across all folds.


For individual feature sets, the highest accuracy (88\%) was achieved using 3D facial landmarks, while the lowest scores were observed for eye gaze (53\%) and pose rotation (56\%) features. Among the combined feature sets based on facial regions, the highest accuracy was achieved for the ``face landmark'' (87\%) region, followed by ``head pose'' (85\%) with just a 2\% difference in accuracy. In contrast, the lowest performance was recorded for facial action units (66\%).

Interestingly, compared to the earlier analysis where all 669 features were used, the 3D landmarks achieved similar performance with just 204 features. Notably, the pose feature set, consisting of only six features, achieved 85\% accuracy, which is just 3\% lower than the highest-performing set. This highlights the potential of reduced feature sets for maintaining strong classification performance while minimizing computational complexity.

\subsubsection{Ablation Study 3} \label{section: AS_binary}
In this analysis, we aimed to understand how the classification model performs in binary classification using facial features for anxiety and non-anxiety detection. We conducted binary classifications between ``Anxious'' and ``Non-Anxious'' by excluding the neutral participants. Similarly, we performed classifications for ``Anxious'' versus ``Neutral'' and ``Neutral'' versus ``Non-Anxious''. Our results showed that the RF model outperformed other classification models. Table \ref{tab:binary_clf_results} presents the classification metrics obtained from binary classification across different feature sets. From Table \ref{tab:binary_clf_results}, we identified several interesting observations that offer insights into comparing multiclass and binary classification and the utility of individual feature categories. First, we found that variations in accuracy and classification metrics in binary classification were consistent with those observed in multiclass classification for different feature sets. For instance, pose rotation features performed poorly across all binary classification cases but showed improved performance when combined with pose location features. This pattern was also observed in multiclass classification. Second, similar to multiclass classification, 3D facial landmarks were the most effective feature set for binary classification. This suggests that facial landmarks are crucial for detecting anxiety and non-anxiety. Third, the binary classification of ``Neutral'' versus ``Non-Anxious'' achieved the best performance, with the highest accuracy of 93\%.

\begin{table*}
    \centering
\caption{Binary classification results of Random Forest averaged over five folds. Abbreviations - Acc., Pr.,  and Re. refer to Accuracy, Precision, and Recall, respectively. Metrics (Acc., Pr., and Re.) can be multiplied by 100 to represent percentages.
 }
\label{tab:binary_clf_results}
    \begin{tabular}{lcccccclcccllllllll} \toprule
   \multirow{2}{*}{\shortstack{\textbf{Feature}  }}& \textbf{Feature Set} & \multicolumn{5}{c}{\textbf{Anxious versus Neutral}} & & \multicolumn{5}{c}{\textbf{Anxious versus non-Anxious}} &&   \multicolumn{5}{c}{\textbf{Neutral versus non-Anxious}}\\ \cmidrule{3-7} \cmidrule{9-13} \cmidrule{15-19}
            
& &  \textbf{\textit{Acc.}}&  \textbf{\textit{Pr.}}&  \textbf{\textit{Re.}}&  \textbf{\textit{F1}}&  \textbf{\textit{AUC}} & &  \textbf{\textit{Acc.}}&  \textbf{\textit{Pr.}}&  \textbf{\textit{Re.}}& \textbf{\textit{F1}}& \textbf{\textit{AUC}} &&   \textbf{\textit{Acc.}}&\textbf{\textit{Pr.}}&\textbf{\textit{Re.}}& \textbf{\textit{F1}}&\textbf{\textit{AUC}}\\ \bottomrule
            \multirow{4}{*}{Eye} 
&gaze
&  0.68&  0.66&  0.62&  0.64&  0.74
 & &  0.65&  0.57&  0.49& 0.52& 0.68
 &&   0.65&0.68&0.7& 0.69&0.73
\\ 
            
&2D landmarks 
&  0.81&  0.8&  0.78&  0.79&  0.91
 & &  0.78&  0.72&  0.72& 0.72& 0.87
 &&   0.76&0.78&0.79& 0.79&0.85
\\ 
            
&3D landmarks
&  0.83&  0.8&  0.81&  0.81&  0.91
 & &  0.86&  0.84&  0.79& 0.81& 0.93
 &&   0.82&0.83&0.85& 0.84&0.9
\\ \cmidrule{2-2}
            
&Combined
&  0.85&  0.85&  0.82&  0.83&  0.94
 & &  0.86&  0.84&  0.81& 0.82& 0.94
 &&   0.85&0.87&0.85& 0.86&0.93
\\ \midrule
            \multirow{3}{*}{\shortstack{Head \\ Pose}} 
&location 
&  0.88&  0.87&  0.85&  0.86&  0.95
 & &  0.87&  0.85&  0.84& 0.84& 0.94
 &&   0.86&0.87&0.88& 0.88&0.93
\\ 
            
&rotation
&  0.72&  0.69&  0.69&  0.69&  0.78
 & &  0.7&  0.66&  0.55& 0.59& 0.75
 &&   0.67&0.7&0.72& 0.71&0.73
\\ \cmidrule{2-2}
            
&Combined
&  0.91&  0.92&  0.87&  0.89&  0.98
 & &  0.88&  0.88&  0.83& 0.85& 0.96
 &&   0.9&0.9&0.91& 0.91&0.97
\\ \midrule
            \multirow{3}{*}{\shortstack{Face \\ Landmark}}
&2D landmarks 
&  0.88&  0.88&  0.84&  0.86&  0.96
 & &  0.86&  0.82&  0.82& 0.82& 0.94
 &&   0.86&0.88&0.86& 0.87&0.94
\\ 
            
&3D landmarks
&  0.91&  0.9&  0.89&  0.9&  0.97
 & &  0.9&  0.89&  0.87& 0.88& 0.97
 &&   0.92&0.91&0.94& 0.93&0.97
\\ \cmidrule{2-2}
   
&Combined
& 0.91& 0.91& 0.89& 0.9& 0.97
 & & 0.9& 0.89& 0.86& 0.87& 0.97
 &&   0.92&0.92&0.94& 0.93&0.97
\\  \midrule
   \multirow{3}{*}{\shortstack{Facial \\ Action \\ Units}}
&intensity
& 0.73& 0.71& 0.67& 0.69& 0.8
 & & 0.72& 0.71& 0.51& 0.59& 0.8
 &&   0.66&0.68&0.75& 0.71&0.74
\\ 
   
&presence
& 0.73& 0.73& 0.65& 0.69& 0.79
 & & 0.72& 0.72& 0.5& 0.59& 0.79
 &&   0.73&0.73&0.81& 0.77&0.77
\\ \cmidrule{2-2}
   &Combined
& 0.79& 0.8& 0.71& 0.75& 0.87
 & & 0.74& 0.76& 0.5& 0.6& 0.85
 &&   0.73&0.73&0.83& 0.77&0.82
\\ \midrule \midrule
 \multirow{3}{*}{ALL} & Top 10\% 
& 0.91& 0.92& 0.87& 0.9& 0.98 & & 0.9& 0.89& 0.86& 0.87& 0.97 && 0.92& 0.93& 0.93& 0.93&0.97\\
   &Top 20\%& 0.92& 0.93& 0.9& 0.91& 0.98
 & & 0.92& 0.91& 0.88& 0.89& 0.98
 && 0.93& 0.94& 0.94& 0.94&0.98
\\ 
 & Top 20\%& 0.92& 0.92& 0.9& 0.91& 0.98 & & 0.91& 0.9& 0.89& 0.89& 0.98 && 0.93& 0.94& 0.94& 0.94&0.98\\ \bottomrule
    \end{tabular}
\end{table*}

\subsection{Feature Importance in Anxiety Classification}  \label{section: feature_imp}
In this work, we use facial video features for anxiety detection, and it is important to understand which facial features are linked to anxiety. To explain the results, we conducted a post-hoc analysis using SHapley Additive exPlanations (SHAP) \cite{lundberg2017unified} to examine the classification model. SHAP quantifies the contribution of each feature to the model's prediction using game-theoretic Shapley values. Additionally, the Shapley values provide insights into how each feature affects the model's decision-making. We will use a Random Forest model trained on the full feature set to identify the important features.

\subsubsection{Multiclass}
\begin{figure}
    \centering
    \includegraphics[width=1\linewidth]{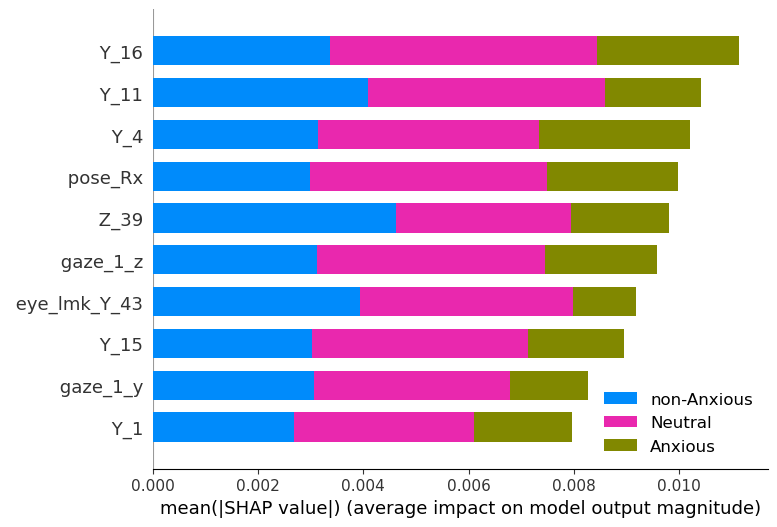}
    \caption{Top ten important features for multiclass classification.
    Face edge (Y\_1, Y\_4, Y\_11, Y\_15, Y\_16, Z\_39), right eye (eye\_lmk\_Y\_43, gaze\_1\_y, gaze\_1\_z), and head position angle (pose\_Rx).
    Best viewed in color.}
    \Description{}
    \label{fig:3_class_shape_summary}
\end{figure}

\begin{figure*}
  \centering
  \begin{subfigure}{0.3\textwidth}
    \centering
    \includegraphics[scale=0.22]{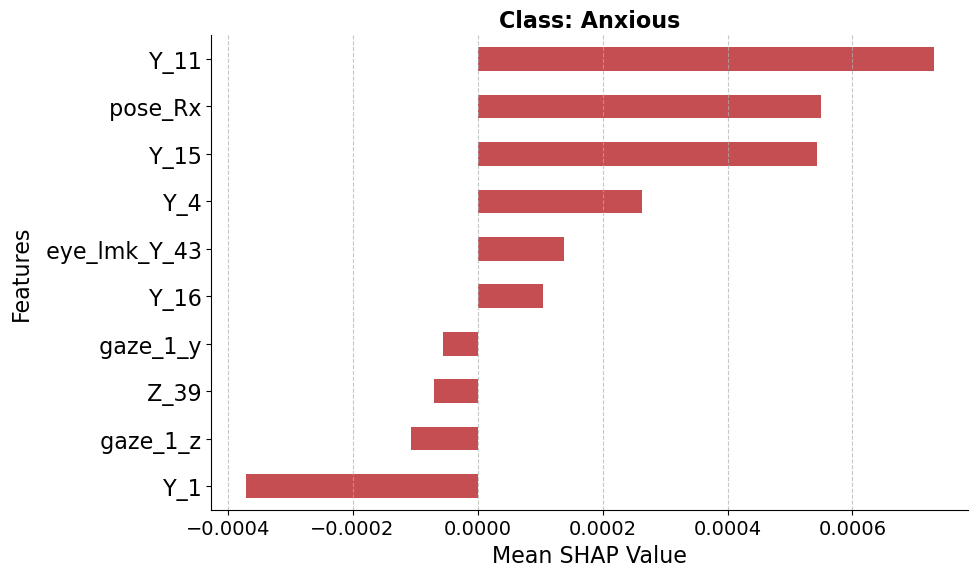}
    \caption{Anxious}
    \Description{}
    \label{fig:3_same_features_Anxious}
  \end{subfigure}
  \begin{subfigure}{0.3\textwidth}
    \centering
    \includegraphics[scale=0.22]{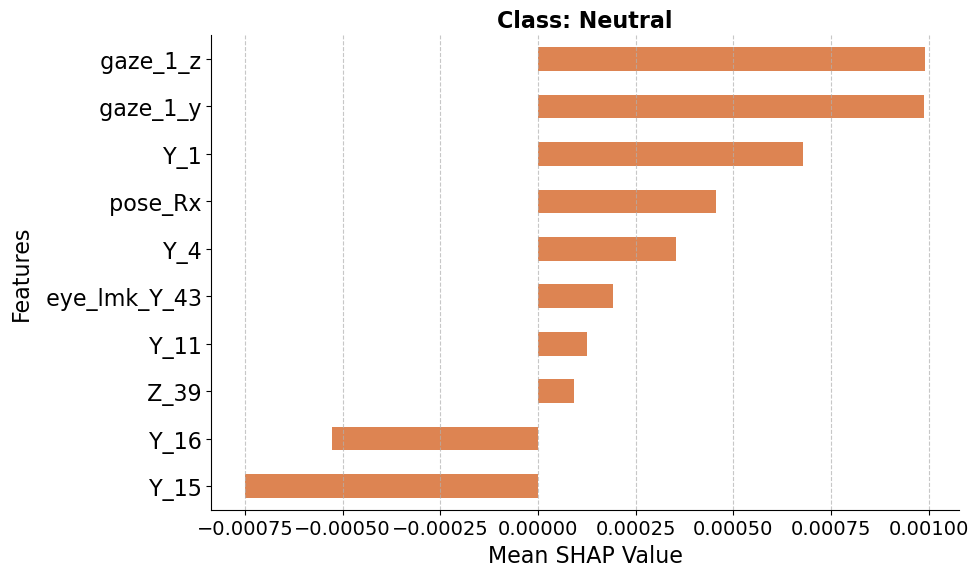}
    \caption{Neutral}
    \Description{}
    \label{fig:3_same_features_Neutral}
  \end{subfigure}
  \begin{subfigure}{0.3\textwidth}
    \centering
    \includegraphics[scale=0.22]{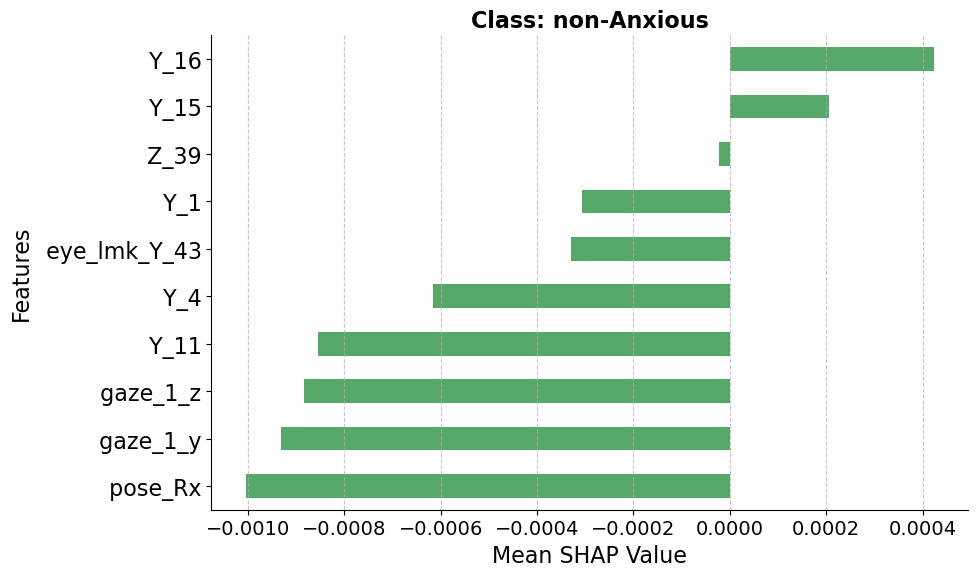}
    \caption{non-Anxious}
    \Description{}
    \label{fig:3_same_features_non-Anxious}
  \end{subfigure}
  \caption{Influence of top ten features from Figure \ref{fig:3_class_shape_summary} on each class in Multiclass classification.}
  \label{fig: 3_same_features_Direction}
\end{figure*}

Figure \ref{fig:3_class_shape_summary} shows the top ten important features for multiclass classification. We observe that features like the face edge (Y\_1, Y\_4, Y\_11, Y\_15, Y\_16, Z\_39), right eye (eye\_lmk\_Y\_43, gaze\_1\_y, gaze\_1\_z), and head position angle (pose\_Rx) are significant for multiclass classification. Further, Figure \ref{fig: 3_same_features_Direction} highlights how these top ten features from Figure \ref{fig:3_class_shape_summary} influence the classification for each class (Figure \ref{fig:3_same_features_Anxious} for anxious, Figure \ref{fig:3_same_features_Neutral} for neutral, and Figure \ref{fig:3_same_features_non-Anxious} for non-anxious). From Figure \ref{fig:3_same_features_Anxious}, we notice that larger values of the face edge (Y\_11, Y\_16) and larger values of head position pitch (pose\_Rx) push the model towards predicting the anxious class.

\begin{figure*}
  \centering
  \begin{subfigure}{0.3\textwidth}
    \centering
    \includegraphics[scale=0.22]{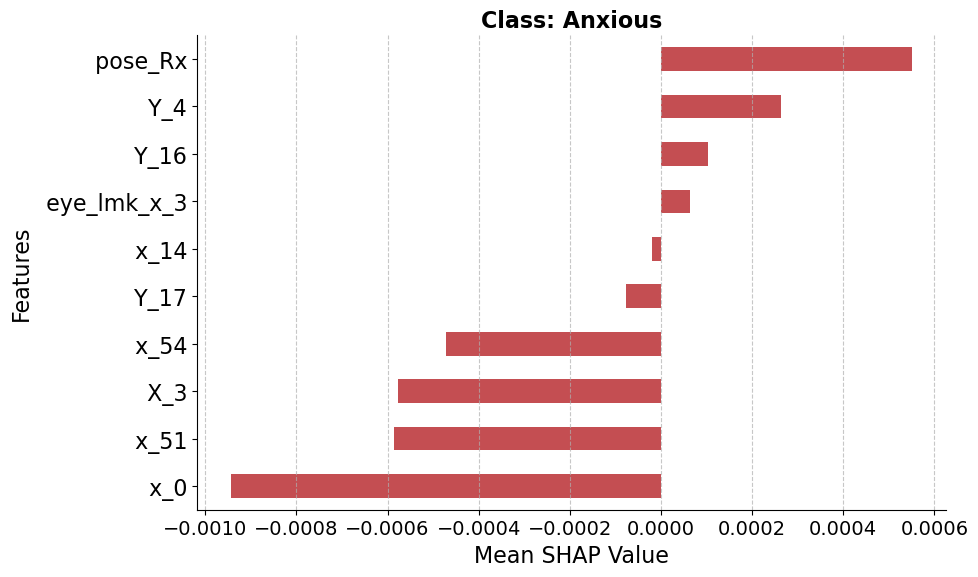}
    \caption{Anxious}
    \Description{}
    \label{fig:3_class_Anxious}
  \end{subfigure}
  \begin{subfigure}{0.3\textwidth}
    \centering
    \includegraphics[scale=0.22]{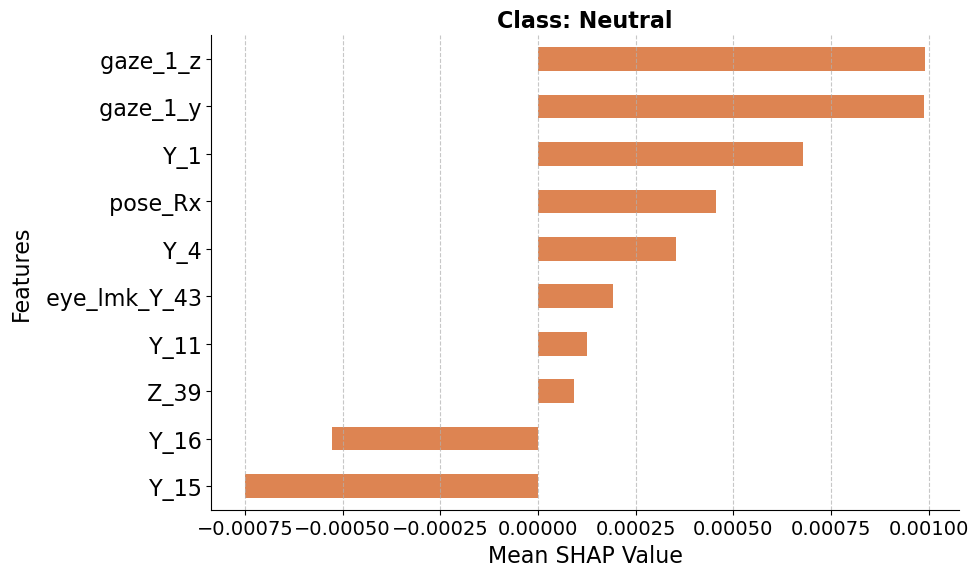}
    \caption{Neutral}
    \Description{}
    \label{fig:3_class_Neutral}
  \end{subfigure}
  \begin{subfigure}{0.3\textwidth}
    \centering
    \includegraphics[scale=0.22]{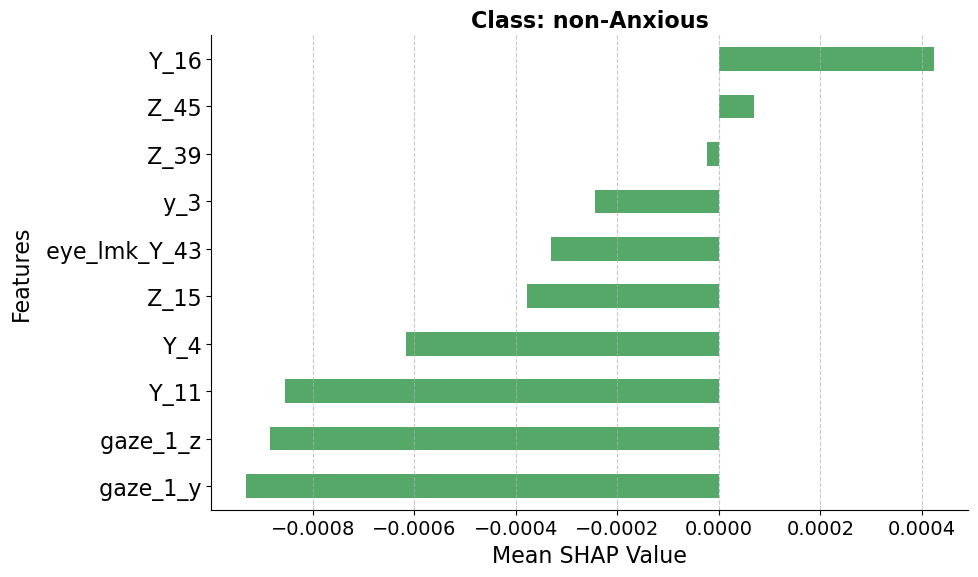}
    \caption{non-Anxious}
    \Description{}
    \label{fig:3_class_non-Anxious}
  \end{subfigure}
  \caption{Influence of top ten features on individual classes in Multiclass classification.}
  \label{fig: 3_class_feature_Direction}
\end{figure*}

Figure \ref{fig: 3_class_feature_Direction} presents the top ten important features of individual classes that influence the model's predictions. For instance, we observe that larger values of head position pitch (pose\_Rx), face edge (Y\_4, Y\_16), and left eye landmark (x\_3) push the model towards predicting the anxious class. On the other hand, smaller values of the remaining features (see Figure \ref{fig:3_class_Anxious}) push the model away from predicting the anxious class. Similarly, we find that larger values of right eye gaze direction (gaze\_1\_z, gaze\_1\_y) and left cheek face contour (Y\_1, Y\_4) push the model towards predicting the neutral class (see Figure \ref{fig:3_class_Neutral}). For the non-anxious class, smaller values of most features (see Figure \ref{fig:3_class_non-Anxious}) push the model away from predicting the non-anxious class.

\subsubsection{Binary Class}
\begin{figure*}
  \centering

  \begin{subfigure}{0.3\textwidth}
    \centering
    \includegraphics[scale=0.25]{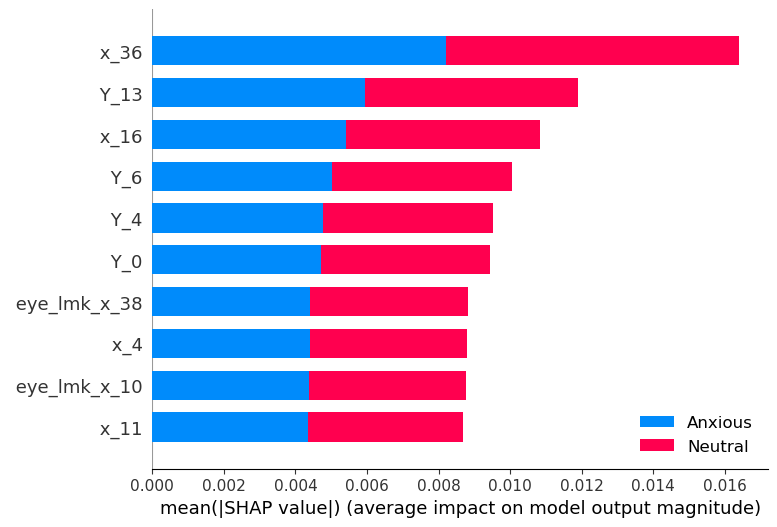}
    \caption{Anxious versus Neutral}
    \Description{}
    \label{fig:2_anxious_neutral_shape_summary}
  \end{subfigure}
  \hfill
  \begin{subfigure}{0.3\textwidth}
    \centering
    \includegraphics[scale=0.25]{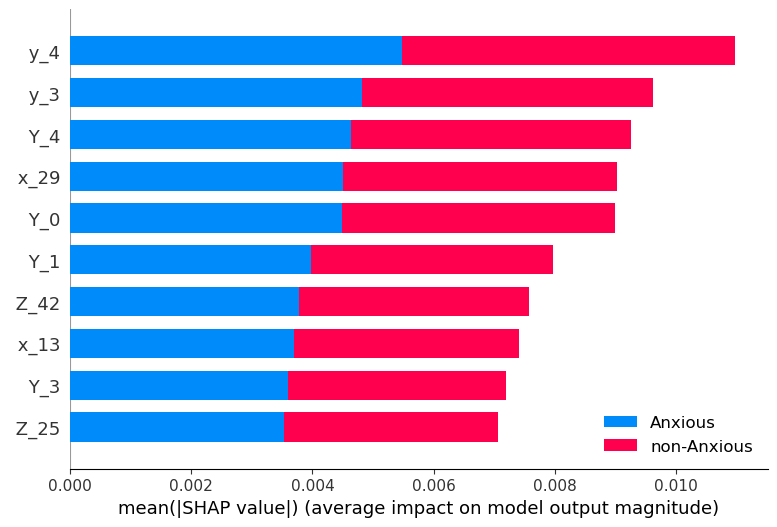}
    \caption{Anxious versus non-Anxious}
    \Description{}
    \label{fig:2_anxious_non-anxious_shape_summary}
  \end{subfigure}
  \hfill
  \begin{subfigure}{0.3\textwidth}
    \centering
    \includegraphics[scale=0.25]{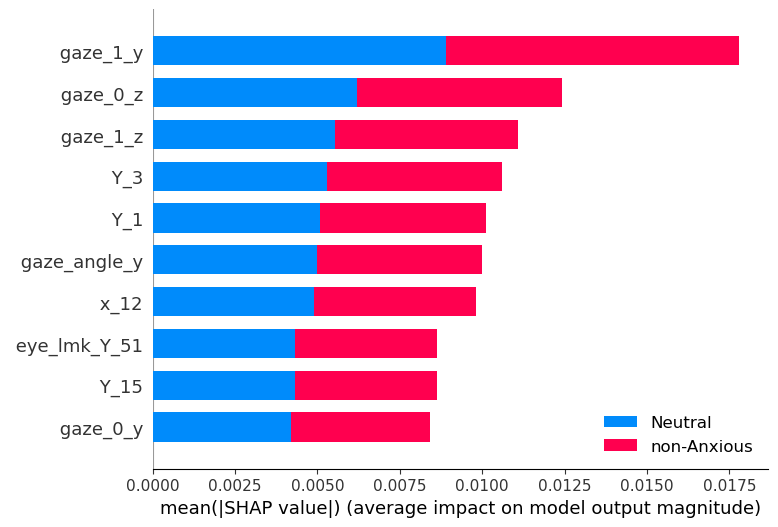}
    \caption{Neutral versus non-Anxious}
    \Description{}
    \label{fig:2_neutral_non-anxious_shape_summary}
  \end{subfigure}

  \vspace{1em} 

  \begin{subfigure}{0.3\textwidth}
    \centering
    \includegraphics[scale=0.22]{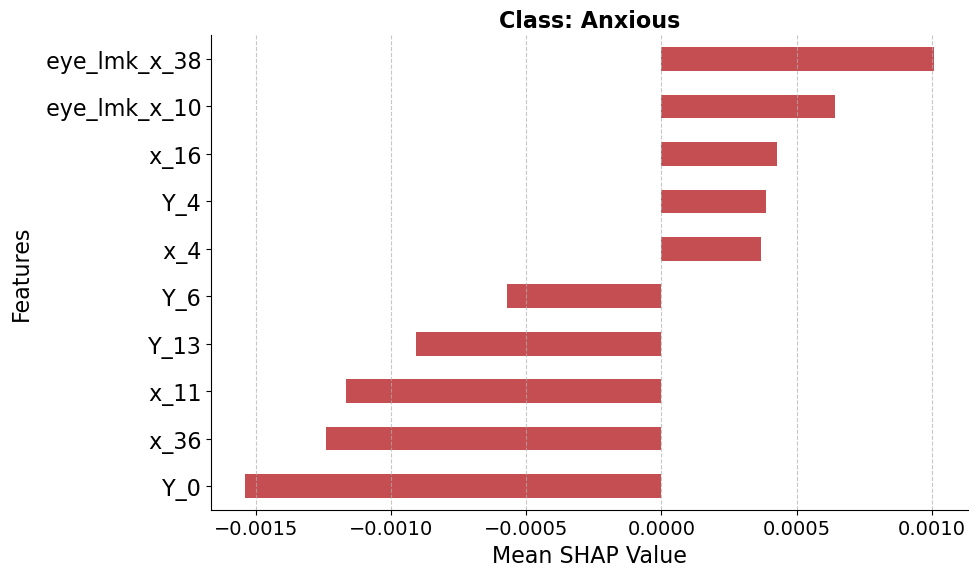}
    \caption{Anxious versus Neutral}
    \Description{}
    \label{fig:2_A_vs_Ne_Anxious}
  \end{subfigure}
  \hfill
  \begin{subfigure}{0.3\textwidth}
    \centering
    \includegraphics[scale=0.22]{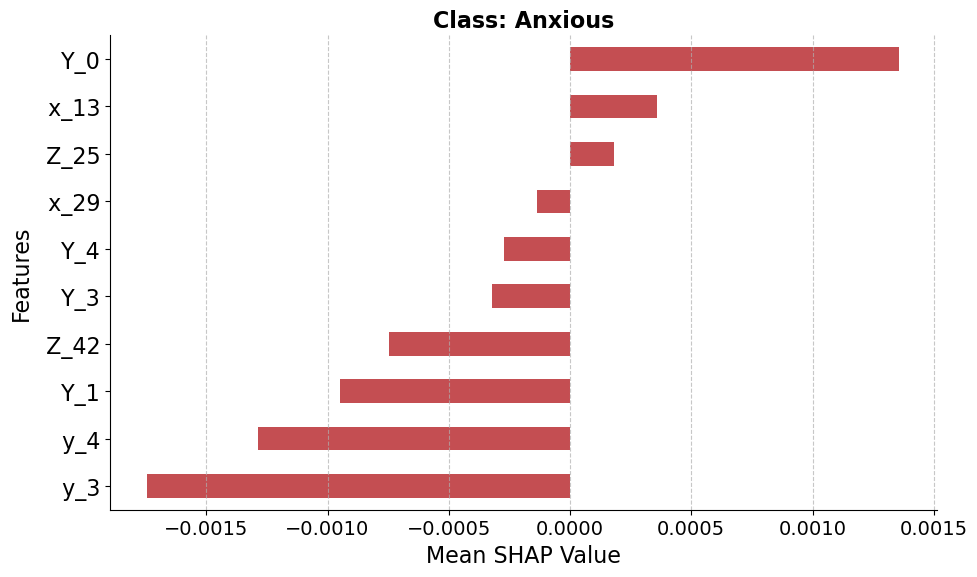}
    \caption{Anxious versus non-Anxious}
    \Description{}
    \label{fig:2_A_vs_NA_Anxious}
  \end{subfigure}
  \hfill
  \begin{subfigure}{0.3\textwidth}
    \centering
    \includegraphics[scale=0.22]{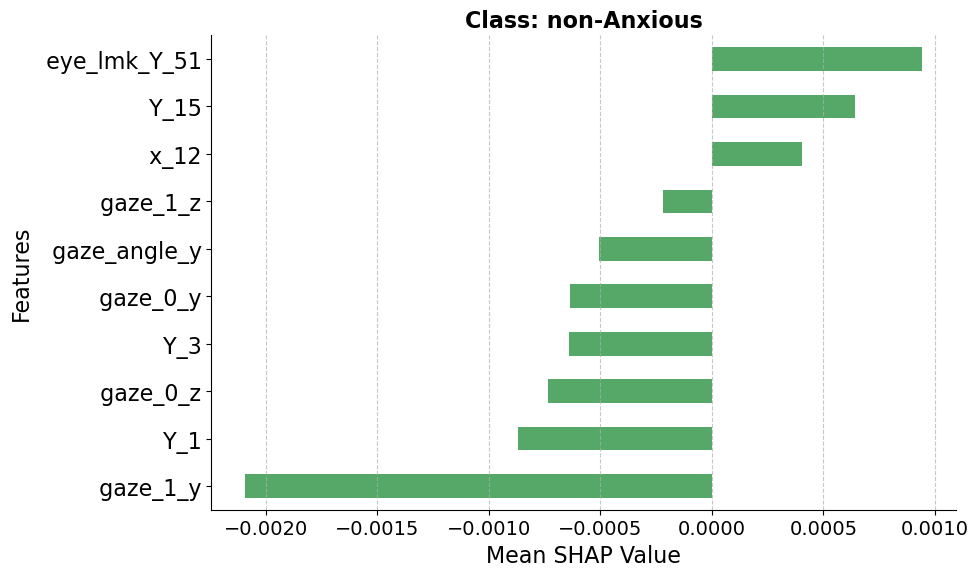}
    \caption{Neutral versus non-Anxious}
    \Description{}
    \label{fig:2_Ne_vs_NA_non-Anxious}
  \end{subfigure}

  \caption{Binary classification SHAP explanations. Figures a, b, and c show the top ten important features. Figures d, e, and f show the influence of features shown in Figures a, b, and c, respectively. Best viewed in color. }
  \label{fig:all_images}
\end{figure*}

Figures \ref{fig:2_anxious_neutral_shape_summary}, \ref{fig:2_anxious_non-anxious_shape_summary}, and \ref{fig:2_neutral_non-anxious_shape_summary} show the top ten features for binary classification in three scenarios: anxious versus neutral, anxious versus non-anxious, and neutral versus non-anxious respectively. We observe that face edges contribute the most to classifications involving anxious versus non-anxious and anxious versus neutral. In contrast, eye-related features dominate in the classification of neutral versus non-anxious.
Figures \ref{fig:2_A_vs_Ne_Anxious}, \ref{fig:2_Ne_vs_NA_non-Anxious}, and \ref{fig:2_Ne_vs_NA_non-Anxious} illustrate the direction of influence for the features shown in Figures \ref{fig:2_anxious_neutral_shape_summary}, \ref{fig:2_anxious_non-anxious_shape_summary}, and \ref{fig:2_neutral_non-anxious_shape_summary} respectively. Figure \ref{fig:2_A_vs_Ne_Anxious} shows the influence of features on the anxious class in the anxious versus neutral classification. We observe that larger values of eye landmarks (i.e., eye\_lmk\_x\_38, eye\_lmk\_x\_10) and larger face edge values (i.e., x\_16, Y\_4, x\_4) push the model towards predicting the anxious class. Conversely, smaller values of other facial landmarks steer the model away from the anxious class, thereby favoring the neutral class. Figure \ref{fig:2_A_vs_NA_Anxious} shows the influence of features for the anxious class in the anxious versus non-anxious classification. Larger values of facial landmarks, such as facial edge features (i.e., Y\_0, x\_13) and the right eyebrow (i.e., Z\_25), push the model towards predicting the anxious class. In contrast, higher values of the remaining facial landmarks influence the model toward predicting the non-anxious class. Figure \ref{fig:2_Ne_vs_NA_non-Anxious} focuses on the influence of features for the non-anxious class in the neutral versus non-anxious classification. Larger values of eye landmarks (i.e., eye\_lmk\_Y\_51) and facial landmarks (i.e., Y\_15, x\_12) push the model towards predicting the non-anxious class. However, smaller values of eye gaze direction for the left and right eyes (i.e., gaze\_0\_y, gaze\_0\_z, gaze\_1\_y, gaze\_1\_z), smaller gaze angles (i.e., gaze\_angle\_y - looking up and down), and smaller left upper cheek values (i.e., Y\_1, Y\_3) steer the model away from the non-anxious class, favoring the neutral class instead.

\subsection{Bias Investigation}  \label{section: bias}
Literature suggests that ML models can sometimes be biased due to factors such as gender, age, etc \cite{cheong2023towards, chu2023age}. Moreover, in \textit{AnxietyFaceTrack}, we use facial features, which can vary based on gender and age \cite{bannister2022sex}. This highlights the need to assess biases in Random Forest model related to these factors. To investigate potential bias in our trained model, we checked if \textit{AnxietyFaceTrack} suffers from any bias. We split our test data into two gender groups: Male and Female. For age, we categorized the test data into Undergrad and Graduate. Additionally, we examined bias based on participants' home locations by dividing the test data into Rural and Urban groups.

\begin{figure*}
  \centering
  \begin{subfigure}{0.3\textwidth}
    \centering
    \includegraphics[scale=0.42]{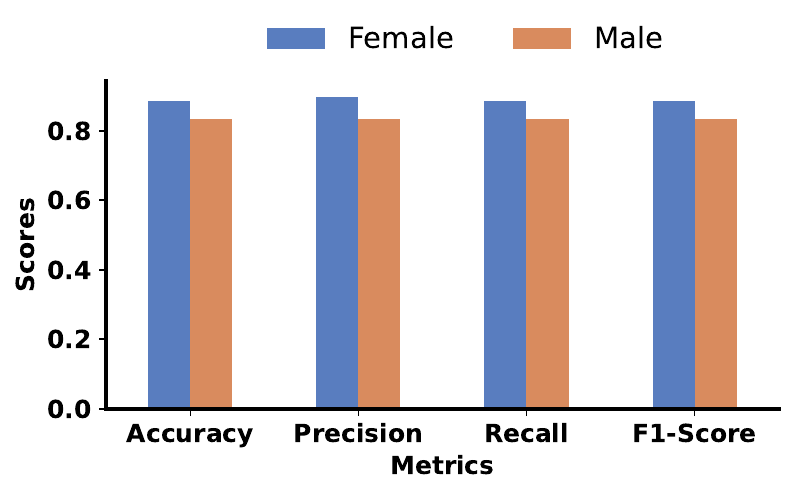}
    \caption{Gender}
    \Description{}
    \label{fig:bias_gender}
  \end{subfigure}
  \begin{subfigure}{0.3\textwidth}
    \centering
    \includegraphics[scale=0.42]{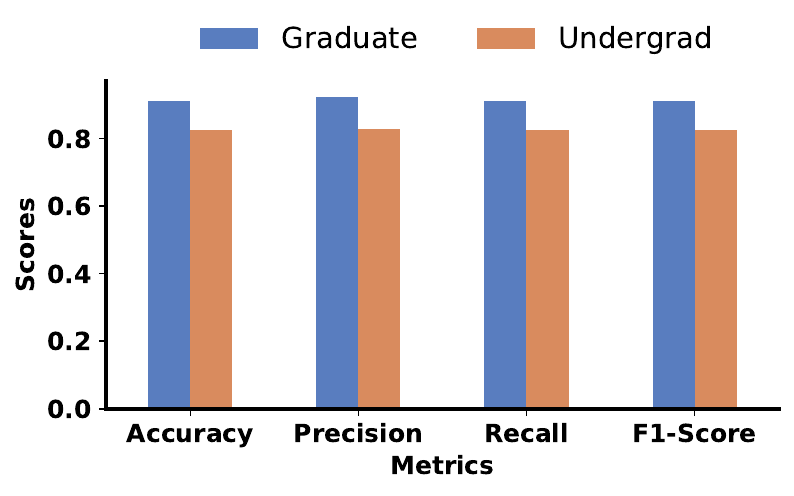}
    \caption{Education Level}
    \Description{}
    \label{fig:bias_age}
  \end{subfigure}
  \begin{subfigure}{0.3\textwidth}
    \centering
    \includegraphics[scale=0.42]{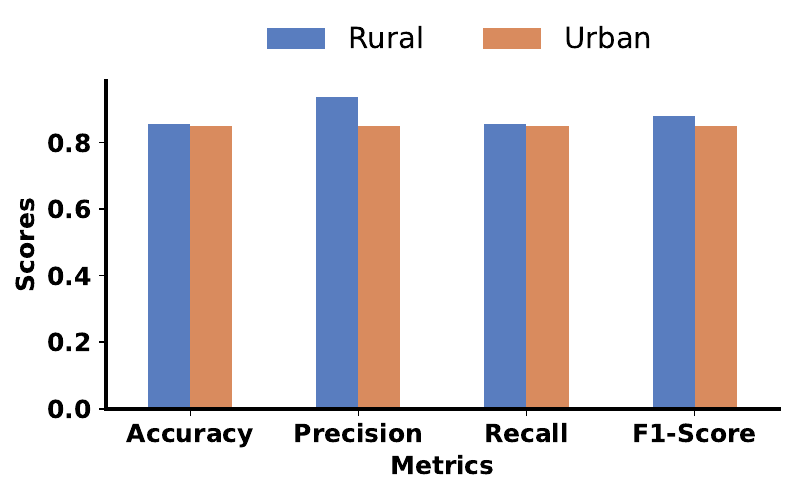}
    \caption{Location}
    \Description{}
    \label{fig:bias_location}
  \end{subfigure}
  \caption{Multiclass classification results while considering (a)
Gender - male and female, (b) Education level - graduate and undergraduate, and (c) Home location - rural and urban. Best viewed in color.}
  \label{fig: bias_investigation}
\end{figure*}

Figure \ref{fig: bias_investigation} shows the performance results of our Random Forest model, revealing several key observations. First, for gender (see Figure \ref{fig:bias_gender}), the classification metrics were higher for females, with a difference of 5-7\% compared to males. This suggests that \textit{AnxietyFaceTrack} works better for females, possibly because females tend to display emotions more prominently through facial expressions than males \cite{parkins2012gender, fischer2015drives, kring1998sex}. Second, for education level (see Figure \ref{fig:bias_age}), we observed about a 10\% difference in classification metrics between graduate and undergrad participants, indicating that the model was better at learning the more subtle behaviors of graduate participants. Lastly, for location (see Figure \ref{fig:bias_location}), the model performed similarly for both rural and urban groups. However, precision was higher for the rural group. Upon closer inspection of the precision for individual labels, we found that precision for rural participants was above 95\%, while for urban participants, it was around 85\%. Our analysis of these biases aims to increase transparency in ML models for anxiety detection and provides valuable insights for future research on anxiety detection.

\section{Discussion}  \label{section: discuss}

In this work, we designed a study, AnxietyFaceTrack, to detect anxiety in participants using facial videos recorded during a social scenario. The study aims to contribute to developing unobtrusive mental health assessment tools. The results of our AnxietyFaceTrack study provide valuable insights into anxiety detection using facial features and machine learning models. Our analysis shows that the Random Forest model, trained on either all 669 features or only the 3D landmark features (\#204), achieved the best overall classification performance, with an accuracy of 88\%. Furthermore, in the ablation study, we found that using just the top 20\% of important features determined by Random Forest feature importance yielded the highest accuracy of 91\% for multiclass classification. Interestingly, we observed that using only six head pose features (i.e., location and rotation) achieved an average accuracy of 85\%, which was just 6\% lower than the best-performing feature set in correctly identifying anxious, neutral, and non-anxious states. For binary classification, the Random Forest model also performed best, achieving average accuracies of 92\% for anxious versus neutral, 91\% for anxious versus non-anxious, and 93\% for neutral versus non-anxious. In summary, the multiclass and binary classification metrics of anxious, neutral, and non-anxious states are promising. Thus highlighting the potential of using facial features for accurate anxiety detection.

Using post hoc analysis, we obtained several key insights. First, we found that anxiety detection in both multiclass and binary classifications achieved similar performance metrics, suggesting that anxious participants can be effectively distinguished from non-anxious and neutral individuals when placed in socially anxious situations. Second, eye gaze performed the worst in both multiclass and binary classifications, even though it has been observed that socially anxious individuals tend to show avoidance gaze behavior compared to non-anxious individuals \cite{schneier2011fear, chen2023social}. Third, individual feature sets struggled in multiclass classification but performed better in binary classification. For example, the action unit feature achieved only 59\% accuracy in multiclass classification, but in binary classification, it achieved an average of 73\% accuracy. This suggests there may be some overlap in facial expressions or movements between anxious, non-anxious, and neutral participants. Fourth, the 3D landmarks and the combined head pose feature set performed the best in both multiclass and binary classification tasks.

Post hoc analysis using SHAP plots provided several insights. For example, facial edges (i.e., facial landmarks 0, 1, 3, 4, 11, 12, 13, 15, 16) and eye landmarks were identified as important features in anxiety detection. Larger values of these features influenced the classification model toward the anxious class. Interestingly, eye gaze features favored the neutral class in both multiclass and binary classifications. Furthermore, similar to the findings of Nepal et al. \cite{nepal2024moodcapture}, we found that 3D landmarks and head pose performed the best compared to other feature sets for anxiety detection. This alignment suggests that these feature sets could be used to develop mental disorder assessment tools.

Furthermore, our investigation into biases in the anxiety classification model revealed several insights. Looking at the classification metrics, we found that the classification model performed better for female and graduate participants. It is important to note that the number of female participants was 36.26\%, and graduate students made up 28.57\%. Despite these proportions, the model was able to learn discriminating patterns more effectively for female and graduate participants. However, upon closer examination, we found that the precision of the anxious class was low for both male and female participants. The recall for the female anxious class was higher compared to males, suggesting that the model was better at capturing the anxious patterns in female participants. Another key finding was that the results were similar for rural and urban participants, even though rural participants made up only 23.08\%. This suggests that, regardless of whether participants grew up in rural or urban areas, the model does not discriminate in favor of one group over the other. These findings on biases provide crucial insights for future research, particularly involving face features and machine learning in anxiety detection.

In conclusion, our \textit{AnxietyFaceTrack} study, which uses facial features extracted from low-cost smartphone camera videos and machine learning for anxiety detection, shows promise as an unobtrusive and continuous approach to mental health assessment, specifically for detecting anxiety.

\subsection{Implications}
Early detection is crucial for enabling timely interventions and promoting recovery for mental disorders \cite{colizzi2020prevention}. This study used facial videos captured through a low-cost smartphone camera for anxiety detection. Our promising results highlight the potential of smartphone-recorded facial videos and machine learning for the early detection of anxiety. This innovative approach can complement existing mental health assessments. Although our study was conducted in a controlled setting, the results pave the way for future research to explore our methodology in real-world settings, leading to a better understanding of anxiety and its early detection in fully naturalistic settings.

Furthermore, our use of low-cost smartphone cameras opens the possibility for anxiety detection through facial features to be feasibly integrated into everyday settings. This technology has the potential to be incorporated into smartphones, enabling early detection and monitoring of anxiety. Extending this work could also assist mental health professionals in routine assessments and interventions, helping to reduce the mental healthcare gap, especially in low-income and developing countries.

\subsection{Limitations}
\textit{AnxietyFaceTrack} study provides valuable insights into an unobtrusive mental health assessment tool for anxiety detection. However, it has certain limitations. First, the study was conducted in a controlled setting, which might limit the study findings in larger settings. However, the findings can serve as a baseline for future research conducted in either controlled or uncontrolled settings for anxiety detection. 
Second, the dataset size is limited due to the small number of participants. However, it is worth noting that machine-learning models were able to identify patterns associated with anxiety. Finally, our study used a self-reported survey questionnaire to create the ground truth, which may be subject to recall bias. Future studies could incorporate multiple questionnaires to reduce potential bias in participants' responses.

\section{Conclusion}  \label{section: conclusion}
Through the \textit{AnxietyFaceTrack} study, we demonstrated the potential of leveraging facial videos recorded using low-cost smartphone cameras and machine learning to detect anxiety in unstaged social settings. Our findings contribute to the growing field of non-intrusive mental health assessment, offering a scalable and accessible solution that seamlessly integrates into everyday smartphone usage.

The study involved 91 participants, with facial videos recorded in a controlled environment simulating a social setting. Facial features extracted from these recordings were used to train both multiclass and binary classification models. The results were promising, with the multiclass model achieving an accuracy of 91\% for distinguishing between anxious, neutral, and non-anxious states. Similarly, the binary classification models achieved accuracies of 92\% for anxious versus neutral, 92\% for anxious versus non-anxious, and 93\% for neutral versus non-anxious comparisons. These outcomes underscore the feasibility of smartphone-based anxiety detection systems and their potential role in advancing personalized mental health care.

\bibliographystyle{ACM-Reference-Format}
\bibliography{sample-base}

\end{document}